\documentclass{article}

\usepackage[preprint, nonatbib]{neurips_2022}

\usepackage[utf8]{inputenc} 
\usepackage[T1]{fontenc}    
\usepackage{hyperref}       
\usepackage{url}            
\usepackage{booktabs}       
\usepackage{amsfonts}       
\usepackage{nicefrac}       
\usepackage{microtype}      
\usepackage{xcolor}         

\usepackage{multirow}
\usepackage{pifont}
\newcommand{\cmark}{\ding{51}}%
\newcommand{\xmark}{\ding{55}}%
\usepackage{subfigure}
\usepackage{graphicx}
\usepackage{amsmath}
\usepackage{url}
\usepackage{booktabs}

\title{Pre-Training a Graph Recurrent Network for Language Representation}

\author{%
  Yile Wang\textsuperscript{\rm 1}\textnormal{,}\ Linyi Yang\textsuperscript{\rm 1}\textnormal{,}\ Zhiyang Teng\textsuperscript{\rm 1}\textnormal{,}\ Ming Zhou\textsuperscript{\rm 2}\textnormal{,}\ Yue Zhang\textsuperscript{\rm 1}  \\
  \textsuperscript{\rm 1}Westlake University\\
  \textsuperscript{\rm 2}Langboat Technology, Beijing, China\\
  \texttt{\{wangyile,yanglinyi,tengzhiyang,zhangyue\}@westlake.edu.cn} \\
  \texttt{zhouming@chuangxin.com} \\
}

\begin{document}

\maketitle

\begin{abstract}
Transformer-based pre-trained models have gained much advance in recent years, becoming one of the most important backbones in natural language processing. Recent work shows that the attention mechanism inside Transformer may not be necessary, both convolutional neural networks and multi-layer perceptron based models have also been investigated as Transformer alternatives. In this paper, we consider a graph recurrent network for language model pre-training, which builds a graph structure for each sequence with local token-level communications, together with a sentence-level representation decoupled from other tokens. The original model performs well in domain-specific text classification under supervised training, however, its potential in learning transfer knowledge by self-supervised 
way has not been fully exploited. We fill this gap by optimizing the architecture and verifying its effectiveness in more general language understanding tasks, for both English and Chinese languages. As for model efficiency, instead of the quadratic complexity in Transformer-based models, our model has linear complexity and performs more efficiently during inference. Moreover, we find that our model can generate more diverse outputs with less contextualized feature redundancy than existing attention-based models.\footnote{We release the code at \url{https://github.com/ylwangy/slstm_pytorch}.}
\end{abstract}

\section{Introduction}

Pre-trained neural models (PTMs)~\cite{gpt2,bert,roberta,xlnet,albert,bart,t5,deberta,electra,gpt3} have been widely used in natural language processing (NLP), benefiting a range of tasks including natural language understanding~\cite{glue,superglue}, question answering~\cite{squad1,squad2}, summarization~\cite{cnndaily,xsum},  and dialogue~\cite{bao-etal-2020-plato,thoppilan2022lamda}. The current dominant methods take the Transformer~\cite{transformer} architecture, a heavily engineered model based on a self-attention network (SAN), showing competitive performance in computation vision~\cite{detr,vit,deit}, speech~\cite{wav2vec} and biological~\cite{AlphaFold2021} tasks. 

Despite its success, Transformer-based models typically suffer from quadratic time complexity~\cite{transformer}, along with the requirement of large computational resources and associated financial and environmental costs~\cite{strubell-etal-2019-energy}. In addition, recent studies show that the attention mechanism, which is the key ingredient of Transformer, may not be necessary~\cite{payless,notallyouneed,synthesizer}. For example, Tay et al.~\cite{synthesizer} find that models learning synthetic attention weights without token-token interactions also achieve competitive performance for certain tasks. Therefore, investigation of Transformer alternatives for pre-trained models is of both theoretical and practical interest. To this end, various non-Transformer PTMs have recently been proposed~\cite{tay-etal-2021-pretrained,gmlp,resmlp,mlpmixer}.

In this paper, we consider a graph neural network (GNN)~\cite{cnn_graph} for language model pre-training. GNN and its variants have been widely used in NLP tasks, including machine translation~\cite{gcn-mt}, information extraction~\cite{graph-event}, and sentiment analysis~\cite{graph-sentiment}. For GNN language modeling, a key problem is how to represent a sentence in a graph structure. From this perspective, ConvSeq2seq~\cite{convseq} can be regarded as a graph convolutional network (GCN)~\cite{gcn} with node connections inside a local kernel. Transformer-based models can be regarded as a graph attention network (GAT)~\cite{gat} with a full node connection.  However, graph recurrent network (GRN)~\cite{slstm,song-etal-2018-graph} models have been relatively little considered.

\begin{table*}[t]
	\centering
	\small
	\begin{tabular}{c|c|c|c|c|c}
	    \hline
    	\bf{Type} & \bf{Models}& \bf{Basic Unit} & \bf{Complexity} & \bf{Parallel}& \bf{Parameter Sharing}\\
    	\hline
    	 \multirow{2}*{LSTM} & Context2Vec~\cite{context2vec} &  \multirow{2}*{RNN} &  \multirow{2}*{$\mathcal{O}(n)$} &\xmark &\xmark \\
    	 &ELMo~\cite{elmo} &   &  &\xmark &\xmark \\
    	\hline
    	\multirow{8}*{Transformer}& GPT2~\cite{gpt2}  &  \multirow{8}*{SAN} & \multirow{8}*{$\mathcal{O}(n^2)$}& \cmark &\xmark\\
    	&BERT~\cite{bert} &&& \cmark &\xmark\\
    	&RoBERTa~\cite{roberta}   & &  &  \cmark&\xmark \\
    	&XLNet~\cite{xlnet}  &  &  & \cmark &\xmark\\
    	&ALBERT~\cite{albert}   & &  &  \cmark &  \cmark \\
    	&BART~\cite{bart}  &  &  & \cmark &\xmark\\
    	&T5~\cite{t5}   & &  & \cmark&\xmark  \\
    	&DeBERTa~\cite{deberta}  & &  & \cmark &\xmark \\
    	\hline
    	\multirow{3}*{Others}&DynamicConv~\cite{tay-etal-2021-pretrained}  &  CNN  & $\mathcal{O}(n)$ & \cmark&\xmark\\
    	\cline{2-6}
    	&gMLP~\cite{gmlp}  &  MLP  & $\mathcal{O}(n)$ & \cmark&\xmark\\
    	\cline{2-6}
    	&Ours &  GRN  & $\mathcal{O}(n)$ & \cmark&  \cmark \\
    	\hline
	\end{tabular}
	\caption{Overview of existing types of pre-trained models and our proposed model.}
	\label{table:modeltypes}
\end{table*}

We follow the GRN structure of Sentence-state LSTM (S-LSTM)~\cite{slstm}, which represents a sentence using a graph structure by treating each word as a node, together with a sentence state node. State transitions are performed recurrently to allow token nodes to exchange information with their neighbors as well as a sentence-level node. Such architecture has shown some advantages over vanilla bidirectional LSTM in supervised sequence classification tasks. However, compared with existing Transformer-based models, its 
potential for general-purpose language modeling pre-training has not been fully exploited.

As seminal models for pre-trained language modeling, ELMo~\cite{elmo} takes a bi-directional LSTM structure, sharing the gate and cell structures as sentence-state LSTMs. BERT~\cite{bert} uses a Transformer encoder with multiple SAN sub-layers. Compared ELMo, computation for each token in S-LSTM is fully parallel, making it practical for large-scale pre-training. Compared with BERT, our attention-free model gives $\mathcal{O}(n)$ complexity instead of $\mathcal{O}(n^2)$ with respect to sequence length, which is more computationally friendly for long sequences. We optimize the model by exploring the suitable architecture design for pre-training, aiming at learning better transferable task-agnostic knowledge for language representations.

Experimental results show that our model can give a comparable performance on language modeling and language understanding tasks for both English and Chinese languages. During inference, our model can gain 2$\sim$3 times speedup or more for extra long sentences against Transformer-based models. 
Moreover, the outputs of our model are more diverse than Transformer-based PTMs, which may help us understand the non-uniform distribution of contextualized embeddings in PTMs from the viewpoint of the model architecture. 

To our knowledge, we are the first to investigate a graph recurrent network for general language model pre-training, which does not rely on the attention mechanism and has linear computational complexity. Experimental results show that our model can largely reduce the inference time cost without much accuracy loss compared with BERT and its variants, on both English and Chinese languages. Our model can serve as a reference for non-Transformer architecture exploration. 

\section{Related Work}
\textbf{Pre-trained Models}. Our work is related to the language model pre-training~\cite{han2021pre}. A comparison of our work and typical existing PTMs is shown in Table~\ref{table:modeltypes}. In general, existing architectures of PTM can be categorized into three families:

1) \textit{LSTM-based}. Context2Vec~\cite{context2vec} is the first to learn generic context embedding using large plain corpora, which is based on bidirectional LSTM~\cite{lstm}. ELMo~\cite{elmo} further uses deep bidirectional LSTM for language modeling pre-training. Combined with the learned features in deep layers, ELMo shows the strong potential of  contextualized representation and pre-training techniques in modern NLP systems.

2) \textit{Transformer-based}. Vaswani et al.~\cite{transformer} proposed Transformer to overcome the non-parallel computation in RNN using long-range attention mechanism. PTMs based on Transformer has developed rapidly over the past few years, including auto-encoding~\cite{bert,roberta,albert,deberta}, auto-regressive~\cite{gpt2,xlnet}, and sequence-to-sequence~\cite{bart,t5} style Transformer variants. 

3) \textit{Others}. Recent work~\cite{payless,notallyouneed,synthesizer} explores whether attention-free models can also be used for pre-training. Tay et al.~\cite{tay-etal-2021-pretrained} show that CNN-based PTM outperforms its Transformer-based counterpart in certain tasks. Liu et al.~\cite{gmlp} propose an MLP-based alternative without self-attention.
Lee-Thorp et al.~\cite{lee2021fnet} propose FNet which uses an unparameterized Fourier transform to replace the self-attention sublayer. These endeavors show that, within proper design and settings, different architectures can also match the performance of Transformer ones in certain scenarios. Our work is in line, while we conducted more experiments on language modeling and general language understanding tasks for both English and Chinese.


\textbf{Graph Recurrent Network}. 
GRN is a particular case of RNN in which the signals at each node in time are supported on a graph. Compared with other graph neural networks like GCN~\cite{gcn} or GAT~\cite{gat}, the information for each node is aggregated recurrently, combined with spatial gating strategies instead of convolutional or self-attention operation. GRN has
been used to model graph structures in NLP tasks, such as semantic role labeling~\cite{gnn-syntax}, machine translation~\cite{gcn-mt}, text generation~\cite{song-etal-2018-graph} and semantic parsing~\cite{xu-etal-2018-exploiting}. In particular,
Zhang et al.~\cite{slstm} use GRN to represent raw
sentences by building a graph structure of neighboring words and a sentence-level node, showing
the success of graph recurrent structure for sequence labeling and classification tasks against bidirectional LSTM. 

Although GRN performs well competitive to RNN or GCN in certain tasks, no previous study investigate its capability for general language modeling and understanding. We fill the gap by exploring the ways of pre-training GRN and demonstrating its effectiveness and efficiency.

\textbf{Time-efficient Transformers.} Due to the quadratic complexity, it has been a practical question how to train a more computationally efficient Transformer. Reformer~\cite{reformer} use locality-sensitive hashing to replace dot-product attention, changing its complexity from $\mathcal{O}(n^2)$ to $\mathcal{O}$($n$\rm{log}$n$).
Longformer~\cite{longformer} utilizes a local windowed attention with global attention, making it easy to process long documents. Linformer~\cite{linformer} optimizes the self-attention using low-rank projection, reducing the complexity to $\mathcal{O}(n)$. Performer~\cite{performer} uses the kernel feature map to efficiently approximate the softmax attention with theoretically grounded $\mathcal{O}(n)$ complexity. Peng et al.~\cite{peng2021rfa} propose random feature methods to approximate the softmax function. For a comprehensive study of these models, we refer the readers to Tay et al.~\cite{tay2020efficient}.

The above works focus on modifying the attention mechanism in Transformer and verify on specific usage\footnote{For example, Reformer conducted experiments on character-level language modeling and synthetic datasets, Longformer utilizes the sparse attention on long context question and summarization.}. Our model achieves linear complexity with respect to the sequence length. However, it does not rely on Transformer while using a different graph structure to represent text generally.

\section{Model}
\begin{figure*}[!t]
	\centering
	\includegraphics[scale=0.75]{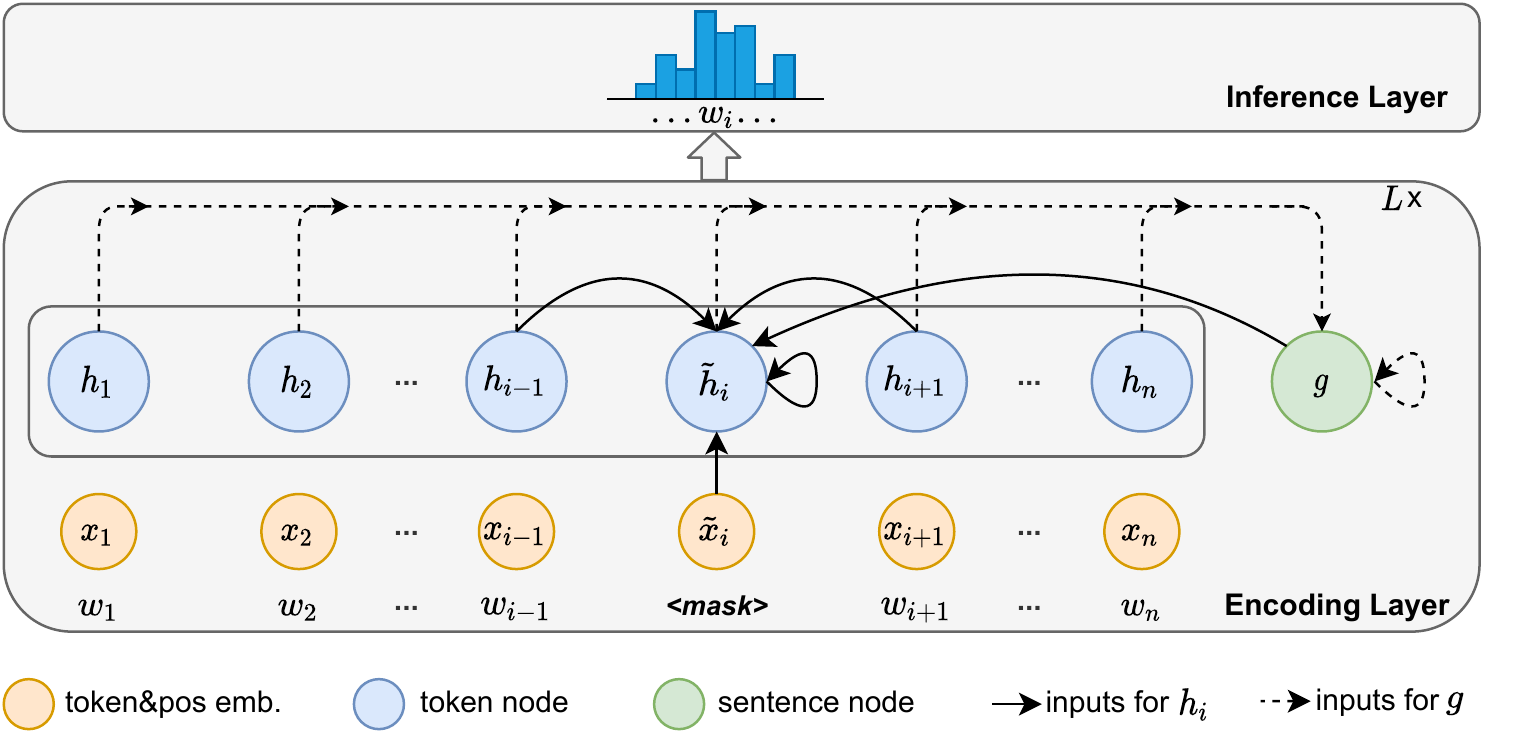}
	\caption{Architecture of our model for language model pre-training. We only show the update of token node $h_i$ and sentence-level node $g$ for brevity. }
	\label{figure:slstm}
	
\end{figure*}

The overall structure of our model is shown in Figure~\ref{figure:slstm}. Following S-LSTM~\cite{slstm}, we treat each sentence as a graph with token nodes and an external sentence state node. The node state is updated in parallel according to the information received in each layer. The representations in the final $L^{th}$ layer is used for inference.
\subsection{Graph Structure}

We build a graph $\mathcal{G} = (\mathcal{V},\mathcal{E})$ for every sentence $s$, which is a sequence of tokens $w_1,w_2,...,w_n$. The verticals $\mathcal{V}$ consists of token nodes $h_i$ ($i=1,2,...,n$) and a sentence state node $g$. The connections are all bi-directional, in the sense that one forward edge and one backward edge exist between node pairs. For edges $\mathcal{E}$, each token node is connected with it neighbor nodes and the sentence node.  Formally,
\begin{equation}
\begin{array}{l}
\mathcal{V} = \{h_1, h_2,..., h_n, g \}\\
\mathcal{E} = \{e_{i,j}, e_{i,g}, e_{g,i}; |i-j| \le 1 \} ,
\end{array}
\label{equation:graph}
\end{equation}
where $e_{i,j}$ is the directed edge from node $h_i$ to $h_j$, $e_{i,g}$ and $e_{g,i}$ are the two directed edges between $h_i$ and $g$. Self-loops are added for all nodes, allowing information flow from its previous state.

\subsection{Encoding Layer}
We strictly follow S-LSTM in node communication. However, different from the original model which uses character-level word representation, we use subwords~\cite{sennrich-etal-2016-neural} and position embeddings~\cite{transformer}. Layer normalization~\cite{ba2016layer} operations are also added, which shows benefit for language model pre-training.

\textbf{Input Embeddings}.
To learn a better vocabulary, we tokenize the text with the language independent SentencePiece~\cite{sentencepiece}. The total size of the vocabulary is 30,000. We first transform each token $w_i$ into token embedding using the trainable lookup table $E$ and the position embedding lookup table $P$, the model input $x_i$ is constructed by summing the two:
\begin{equation}
x_i = E(w_i) + P(w_i)
\label{eq:eq2}
\end{equation}

\textbf{Token Node Update}.
We initialize hidden states and hidden cells for each token node, the sentence-level node with $h_1^0, h_2^0,...,h_n^0,g^0$ and $c_1^0, c_2^0,...,c_n^0,c_g^0$, respectively. In each layer $t$ ($t=1,2,...,L$), the token node states $h_i^t$ is calculated using gating mechanism similar with LSTM:
\begin{equation}
\begin{array}{l}
\begin{aligned}
&\xi_i^{t-1} = h_{i-1}^{t-1} \mathbin\Vert h_{i}^{t-1} \mathbin\Vert h_{i+1}^{t-1} \\
&\hat{i}_i^t = \sigma({\rm LayerNorm}(W_i\xi_i^{t-1} + U_ix_i + V_ig^{t-1} + b_i)) \\
&\hat{l}_i^t = \sigma({\rm LayerNorm}(W_l\xi_i^{t-1} + U_lx_i + V_lg^{t-1} + b_l)) \\
&\hat{r}_i^t = \sigma({\rm LayerNorm}(W_r\xi_i^{t-1} + U_rx_i + V_rg^{t-1} + b_r)) \\
&\hat{f}_i^t = \sigma({\rm LayerNorm}(W_f\xi_i^{t-1} + U_fx_i + V_fg^{t-1} + b_f))\\
&\hat{s}_i^t = \sigma({\rm LayerNorm}(W_s\xi_i^{t-1} + U_sx_i + V_sg^{t-1} + b_s)) \\
&o_i^t = \sigma({\rm LayerNorm}(W_o\xi_i^{t-1} + U_ox_i + V_og^{t-1} + b_o)) \\
&u_i^t = \tanh({\rm LayerNorm}(W_u\xi_i^{t-1} + U_ux_i + V_ug^{t-1} + b_u)) \\
&i_i^t,l_i^t,r_i^t,f_i^t,s_i^t={\rm softmax}(\hat{i}_i^t,\hat{l}_i^t,\hat{r}_i^t,\hat{f}_i^t,\hat{s}_i^t)\\
&c_i^t = l_i^t\odot c_{i-1}^{t-1} + f_i^t\odot c_i^{t-1} + r_i^t\odot c_{i+1}^{t-1} \\
&\qquad\qquad\quad\ \ \ \ \ \ \  + s_i^t\odot c_g^{t-1} + i_i^t\odot u_i^{t}\\
&h_i^t=o_i^t\odot\tanh(c_i^t)
\end{aligned}
\end{array}
\label{eq:eq3}
\end{equation}
where $ \mathbin\Vert$ is concatenation operation, $\xi_i^{t-1}$, $x_i$, $g^{t-1}$ represent the inputs from previous local states, token embedding  and previous global
states, respectively. In Eq.~\ref{eq:eq3}, we calculate multiple LSTM-style gates to control the corresponding information flow. $\hat{l}_i^t,\hat{r}_i^t,\hat{f}_i^t,\hat{s}_i^t$ are the forget gates with respect to the left token cell $c_{i-1}^{t-1}$,  right token cell $c_{i+1}^{t-1}$, 
current token cell $c_{i}^{t-1}$ and sentence cell $c_g^{t-1}$. $\hat{i}_i^t$, $o_i^t $ are the input gate and output gate, respectively. Layer normalization is used to control the distributions of neurons in each gate. $W_x$, $U_x$, $V_x$ and $b_x$ ($x \in \{i,l,r,f,s,o,u\}$) are model parameters.

\textbf{Sentence-level Node Update}.
The sentence-level node $g^t$ takes the previous token state as inputs and is calculated by :
\begin{equation}
\begin{array}{l}
\begin{aligned}
&\overline{h} = {\rm avg}(h_1^{t-1},h_2^{t-1},...,h_n^{t-1}) \\
&\hat{f}_i^t = \sigma({\rm LayerNorm}(W_fg^{t-1}+U_fh_i^{t-1} + b_f)) \\
&\hat{f}_g^t = \sigma({\rm LayerNorm}(W_gg^{t-1}+U_g\overline{h} + b_g)) \\
&o^t = \sigma({\rm LayerNorm}(W_og^{t-1} + U_o\overline{h}+b_o))\\
&f_1^t,...,f_n^t,f_g^t = {\rm softmax}(\hat{f}_1^t,...,\hat{f}_n^t,\hat{f}_g^t)\\
&c_g^t = f_g^t\odot c_g^{t-1} + \sum{f_i^t\odot c_i^{t-1}}\\
&g^t=o^t\odot\tanh(c_g^t)
\end{aligned}
\end{array}
\label{eq:eq4}
\end{equation}
where $\hat{f}_1^t,...,\hat{f}_n^t,\hat{f}_g^t$ are the forget gates with respect to token cells $c_1^{t-1},...,c_n^{t-1}$ and sentence cell $c_g^{t-1}$, $o^t$ is the output gate. $W_x$, $U_x$ and $b_x$ ($x \in \{g,f,o\}$) are model parameters. The generated hidden state $h_i^t$ and $g^t$ are sent to the next layer, together with the memory state $c_i^t$ and $c_g^t$. 

Both LSTM and our model use gates to control the information exchange, the main difference is that LSTM represents the subsequence from the beginning to a certain token in the sequence direction, while our model uses a structural state to represent all tokens and a sentence node simultaneously in the layer direction.

\subsection{Inference Layer}
\textbf{Mask Language Modeling}. The final layer representations are used for inference. In self-supervised learning without labels, similar to the auto-encoding style pre-training in BERT~\cite{bert}, we randomly mask 15\% input tokens and use its corrupted representations for masked language modeling prediction\footnote{Following Liu et al.~\cite{roberta} and Izsak et al.~\cite{izsak-etal-2021-train}, we do not use sentence-level objective such as next sentence prediction (NSP).}, the final training loss is:
\begin{equation}
\begin{array}{l}
\begin{aligned}
\mathcal{L}_{mlm} &= -\sum_{i}{\log P_\theta( {w}_{i}| \widetilde{w}_{i})} \\
&= -\sum_{i}{\log \frac{\exp(E(w_i)^\top (W\widetilde{h}_{i}))}{\sum_{j}{\exp(E(w_j)^\top (W\widetilde{h}_{i}))}}} 
\end{aligned}
\end{array}
\end{equation}
where $\widetilde{w}_{i}$ is the masked token, $\widetilde{h}_{i}$ is the corresponding representation, $W$ is an added linear output layer and $E(w_i)$ is the ground truth token embedding in Eq.~\ref{eq:eq2}.

\begin{figure*}[t]
	\centering
	\includegraphics[scale=0.75]{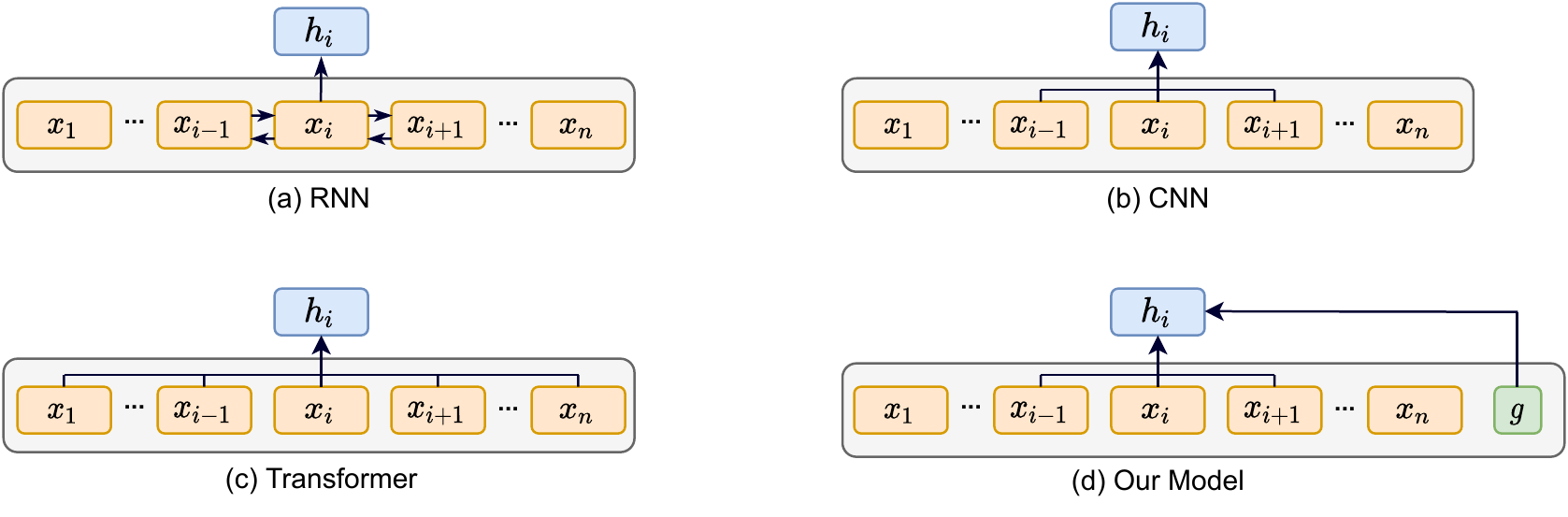}
	\caption{Comparison between different model architectures for hidden states generation in each layer.}
	\label{figure:type}
\end{figure*}

\section{Comparison with Other Models}
Figure~\ref{figure:type} compares the ways of hidden states generations of our model with RNN-based~\cite{context2vec,elmo}, CNN-based~\cite{convseq,tay-etal-2021-pretrained}, and Transformer-based~\cite{bert,roberta,albert} models.

\textbf{Token Representation}. For token node update, our model is similar to CNN in the way of integrating local context, despite that CNN uses different kernels for capturing information from different distances. In a higher layer of both architectures, the reception field for each tokens become larger (i.e, each token can receive long-range context). Different from CNN, we explicitly model sentence-level information as a feature for each token, which provides global information.

\textbf{Sentence Representation}. Transformer-based models (e.g., BERT, RoBERTa) use a special symbol [CLS], attached at the beginning of a sentence, as the sentence representation, which is treated the same to other tokens as model inputs for calculating hidden representation. Recent work~\cite{ke2021rethinking} shows that this forbids the decoupling of sentence representation from token representations, which limits model expressiveness. In our model, the sentence-level node representation is designed to be separated from other tokens, which can benefit the diversity of output vectors.

\textbf{Parameter Sharing}. We make all the trainable parameters in Eq.~\ref{eq:eq3} and Eq.~\ref{eq:eq4} shared across GNN layers, which is similar to the parameters in LSTM along the sequence direction~\cite{elmo}. Layerwise parameter sharing is also used in models such as Universal Transformer~\cite{Universal-Transformers} and ALBERT~\cite{albert}. However, both are based on Transformer architecture, which uses attention to achieve information exchange across all token pairs, while we use graph structure with an LSTM-style gating mechanism for token and sentence-level node update interactively.

\textbf{Complexity}. Transformer-based models update each token by attending over the whole contexts (i.e., all other tokens) in each layer, resulting in a complexity of $\mathcal{O}(n^2)$ w.r.t sequence length. For our model, we update each token using its fixed local context together with a sentence-level representation in each layer, making our model has $\mathcal{O}(n)$ complexity.

\section{Experiments}
We first explore the suitable architecture settings by language modeling task. Then we pre-trained using large-scale corpus and verify the effectiveness of our model for learning transferable knowledge on both English and Chinese NLP tasks.

\subsection{Development Experiments on Language Modeling}
\begin{figure*}[!t]
	\centering
	\includegraphics[scale=0.28]{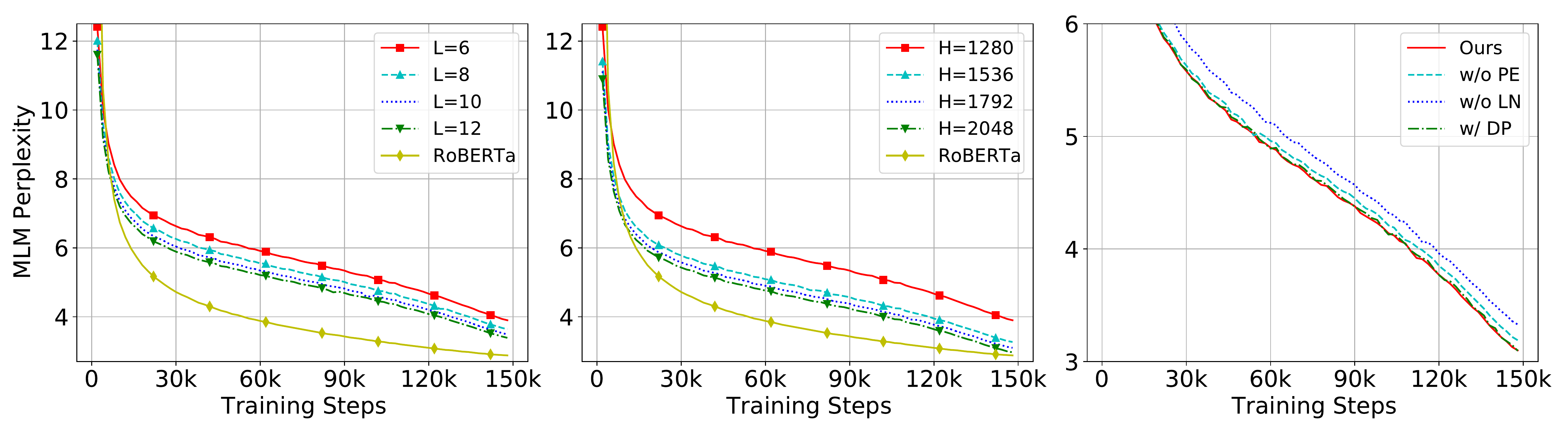}
	\caption{Performance of mask language modeling with different settings. Left(a): layers; Mid(b): hidden size; Right(c): removing position embedding, removing layer normalization and adding dropout in each layer.}
	\label{figure:dev}
\end{figure*}

We train our model in different settings on the WikiText-103 dataset~\cite{wikitext} with 64 batch size and 150k steps, evaluating on the perplexity for predicting the masked token\footnote{Because our model encodes text from both forward and backward direction simultaneously, we consider the cloze-style (masked) language modeling rather than the left-to-right (causal) language modeling.}. We use the RoBERTa-base model for comparison, which is a Transformer based model with 12 layers, 12 attention heads, 768 hidden sizes, and a total of 125 million parameters. The results are shown in Figure~\ref{figure:dev}.

\textbf{Recurrent Step}. 
In Figure~\ref{figure:dev}(a), we first set the dimension as 1280 (with a total of 107 million parameters) and vary the layer number. Empirically, we find that the perplexity decreases steadily while the model becomes deeper. However, the difference is insignificant when the layer number reaches 10 or more.
Also, although the training parameters stay unchanged because of the parameter sharing mechanism, the total account of calculation still increases with more recurrent steps, which makes it 1/1.2/1.5/1.8 times slower when the layer is 6/8/10/12, respectively. Our model generally convergences slower at the beginning, however, the perplexity gap compared with RoBERTa gradually become smaller at the final stage.

Lan et al.~\cite{albert} suggest that there is no need for deeper models when sharing all cross-layer parameters. Zhang et al.~\cite{slstm} also find the accuracy reaches a peak when the recurrent step is around 9. The benefit for very deep recurrent-style models can be negligible while the time cost is higher. The reason can be that, for our model, the top reception field for each token is roughly enough when the recurrent step is around 10.

\textbf{Hidden Size}. 
We then set the layer number as 10 and try different hidden sizes to enhance the model capacity. As shown in Figure~\ref{figure:dev}(b), by adding the hidden size, the model capacity becomes larger and the perplexity decreases much, while the difference becomes small too when the number reaches 1792 or more. Hidden size defines the total parameter size directly and the total training time cost is also influenced, where the model takes 1/1.4/1.8/2.2 times training times when the hidden size is 1280/1536/1792/2048, respectively. We find that the perplexity is almost as small as RoBERTa when the hidden size reaches 1792, which indicates that our model can also give competitive results as Transformer-based models for language modeling.

\textbf{Position Embedding, Layer Normalization, and Dropout}. 
In Figure~\ref{figure:dev}(c), when we remove the position embeddings in Eq.~\ref{eq:eq2}, the perplexity becomes slightly higher, showing that the global position information is useful for our model, although the different local left or right gated information is used for each token. The way of integrating both relative and absolute position information is also similar to DeBERTa~\cite{deberta}, where they add absolute embeddings in the inference layer. 

Layer normalization~\cite{ba2016layer} is used to make distribution stability through normalizing the mean and
variance of all summed inputs to the neurons in one layer, which is
irreplaceable in Transformer-based models. In our model, we also find it is crucial for doing layer normalization to the various gates in Eq.~\ref{eq:eq3} and Eq.~\ref{eq:eq4}, where the perplexity becomes larger without such operation. 

There is no significant improvement when using a dropout rate of 0.1 for output regularization as compared with non-dropout architecture. Lan et al.~\cite{albert} also find that adding dropout in large Transformer-based models may have harmful results. We suppose that in recurrent-style models like ALBERT and ours, the interactions between layers are more strict, so the inconsistent dropout may lose useful learned features during training, which makes it not much useful as in other models.

\subsection{Pre-training}
\textbf{Dataset}. English models are trained using the latest Wikipedia and BookCorpus~\cite{bookcorpus}. Chinese models are trained using Wikipedia. The total amount of training data is on par with BERT~\cite{bert} for both languages.

\textbf{Baselines}. Strictly comparing the PTMs is difficult because of the different dataset processing, training strategies, and environmental settings. As shown in Table~\ref{table:baselines}, we consider the most related and popular models with similar training corpus for comparison. For English, we use the published RNN-based model (ELMo), compact version of BERT (DistilBERT), BERT, and recurrent version of BERT (ALBERT). For Chinese, we add some BERT variants which use Chinese word segmentor for whole word masking~\cite{bertwwm}, or modify the masked token prediction as a correction target~\cite{macbert}.\footnote{We use baseline checkpoints downloaded from \url{https://allenai.org/allennlp/software/elmo} and \url{https://huggingface.co/models}.}

\textbf{Settings}. Similar to BERT, we train with a batch size of 128 and a maximum length of 512 for 300k steps, using Adam optimizer with learning rate $lr$=0.003, $\beta_1$=0.9, $\beta_2$=0.98, learning rate warmup over the first 3,000 steps, and linear decay with 0.03. Both English and Chinese model has 10 layers, and 1792 hidden size, where the total parameter size is 186M. We use 8 NVIDIA GeForce RTX 3090 GPUs for pre-training and it takes around 10 days.  All the pre-training implementations are based on FairSeq~\footnote{\url{https://github.com/pytorch/fairseq}}~\cite{fairseq} framework.

\begin{table}[t]
	\centering
    \small
	\begin{tabular}{l|l|c}
	 \hline
    	 \textbf{Models (English)}&\textbf{Pre-training Data}&\textbf{Objective}\\
    	 \hline
    	 ELMo~\cite{elmo} &1 Billion Word &CLM\\
    	 DistilBERT~\cite{distillbert}&Wiki+BooksCorpus&BERT+KD\\
    	 BERT-base~\cite{bert} &Wiki+BooksCorpus&MLM+NSP\\
    	 ALBERT-base~\cite{albert} &Wiki+BooksCorpus&MLM+SOP\\
	     \hline
    	 \textbf{Models (Chinese)}&\textbf{Data}&\textbf{Objective} \\
    	 \hline
    	 BERT-base~\cite{bert}&Wiki&MLM+NSP\\
    	 BERT-wwm~\cite{bertwwm}&Wiki&MLM+NSP\\
    	 BERT-wwm-ext~\cite{bertwwm}&Wiki+EXT&MLM+NSP\\
    	 RoBERTa-wwm~\cite{cluecorpus}&Wiki+CLUECorpus&MLM\\
    	 RoBERTa-wwm-ext~\cite{bertwwm}&Wiki+EXT&MLM\\
    	 ALBERT-large~\cite{albert}&Wiki+EXT&MLM+SOP\\
    	 MacBERT~\cite{macbert}&Wiki+EXT&Mac+SOP\\
    	\hline
	\end{tabular}
	\caption{Baseline models. CLM: casual language modeling. KD: knowledge distillation. SOP: sentence order prediction. Mac: MLM as correction. wwm: whole word masking. ext: external training data.}
	\label{table:baselines}
\end{table}

\begin{figure*}[t]
	\centering  
	\includegraphics[scale=0.66]{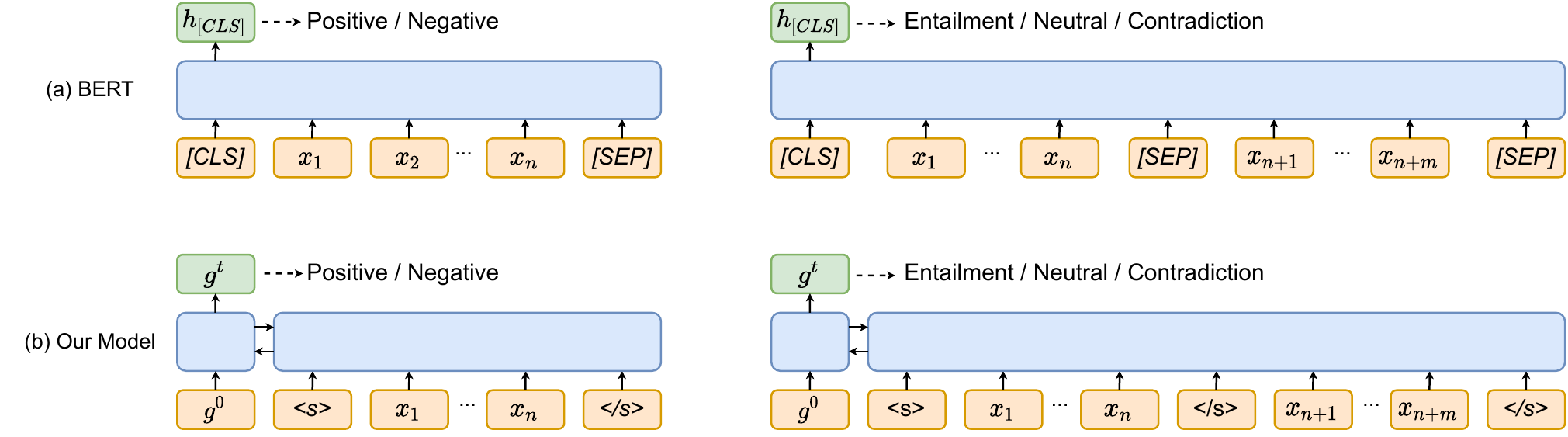}
	\caption{Illustrations of fine-tuning BERT and our model on different tasks. Left: single sentence tasks (e.g., Sentiment Analysis); Right: sentence pair tasks (e.g., Multi-Gerne Natural Language Inference).}
	\label{figure:finetune}
\end{figure*}


\begin{table*}[t]
	\centering
	\small
    \begin{tabular}{lcccccccc}
    \toprule
    \textbf{Model}  & \textbf{CoLA} & \textbf{SST2} & \textbf{MRPC} & \textbf{QQP} & \textbf{MNLI} & \textbf{QNLI}  & \textbf{RTE} & {\textit{Avg}}. \\ 
    \midrule
    ELMo~\cite{elmo}  & 44.1 & 91.5 & 70.8 & 88.0 & 68.6 & 71.2  & 53.4 & 69.65\\
    DistilBERT~\cite{distillbert}  & 51.3 & 91.3 & 82.7 & 88.5 & 82.2 & 89.2  & 59.9 & 77.87 \\
    BERT-base~\cite{bert}  & \textbf{56.3}&\textbf{91.7}  &83.5 & \textbf{89.6} & \textbf{84.0} & \textbf{90.9} & 65.3 & 80.18 \\
    ALBERT-base~\cite{albert}  & 48.2 & 90.7 & \textbf{87.2} & 88.2 & 82.3 & 90.1  & \textbf{69.7} & 79.48 \\
    Ours & 55.3 &90.3 & 81.0 & 88.8 & 81.4 & 89.6  & 64.3 & 78.67\\
    \bottomrule
    \end{tabular}
	\caption{Results on GLUE benchmark dev sets. Scores for CoLA are Matthews correlation, others are reported by the accuracy.}
	\label{table:glue}
\end{table*}

\begin{table*}[t]
	\centering
	\small
	\begin{tabular}{lccccccc}
        \toprule

    	 \textbf{Model} & \textbf{TNEWS} & \textbf{IFLYTEK}  & \textbf{CSL} &\textbf{LCQMC}&\textbf{THUCNews}&\textbf{CSC} & {\textit{Avg}}.\\
    	 \midrule
    	 BERT-base~\cite{bert}&56.14&59.67&81.40&87.89&95.35&92.58&78.83\\
    	 BERT-wwm~\cite{bertwwm}&56.47&59.71&81.23&87.93&95.28&93.00&78.93\\
    	 BERT-wwm-ext$^\dagger$~\cite{bertwwm}&57.35&59.90&80.86&88.05&95.43&93.00&79.09\\
    	 RoBERTa-wwm$^\dagger$~\cite{cluecorpus}&57.29&59.29&81.16&88.41&95.19&93.25&79.09\\
    	 RoBERTa-wwm-ext$^\dagger$~\cite{bertwwm}&57.09&\textbf{60.71}&81.80&88.68&95.69&93.33&79.55\\
    	 ALBERT-large~\cite{albert}&55.69&58.36&80.46&88.27&93.52&91.25&77.92\\
    	 MacBERT$^\dagger$~\cite{macbert}&57.50&59.36&\textbf{81.83}&\textbf{89.18}&\textbf{95.74}&\textbf{93.33}&79.49\\
    	 Ours&\textbf{57.56}&60.10&80.73&86.06&95.17&93.08&78.78\\
    	\bottomrule
	\end{tabular}
	\caption{Results on Chinese tasks dev sets, and scores are reported by the accuracy.}
	\label{table:clue}
\end{table*}

\subsection{Fine-tuning}

\textbf{Evaluating Benchmarks}. For English tasks, we evaluate our pre-trained models on tasks in GLUE~\cite{glue}, including linguistic acceptability (CoLA), sentiment analysis (SST), sentence pair similarity (MRPC, QQP), and natural language inference (MNLI, QNLI, RTE). 

For Chinese tasks, we evaluate on short and long text classification (TNEWS, IFLYTEK), keywords matching (CSL)~\cite{clue}, question matching (LCQMC)~\cite{lcqmc}, document classification (THUCNews)~\cite{thucnews} and sentiment analysis (ChnSentiCorp, CSC for short)~\cite{sa}.

\textbf{Settings}. Although our model architecture is different from most baselines, the fine-tuning strategy for each task can be the same as BERT-style models. As shown in Figure~\ref{figure:finetune}, the output of the sentence node $g$ can be treated as the representation of [CLS] in BERT, which can be used for single sentence classification tasks directly. For sentence pair classification tasks, we concatenated two sentences and the target label is still predicted using the sentence node $g$ in the last layer.

We use the official code from Huggingface~\cite{huggingface} and CLUE~\cite{clue} for reproducing  the baseline and our results without external data augmentation, we mainly tune the parameters with training epochs in \{2, 3, 5, 10\}, learning rate in \{2e-5, 3e-5, 5e-5\} and batch size in \{16, 32, 64\}.

\subsection{Results}

The results are shown in Table~\ref{table:glue} and Table~\ref{table:clue}. For English tasks, our model gives a average score of 78.67, which is higher than ELMo result of 69.65 and on par with Transformer-based baselines (77.87$\sim$80.18). Compared with Transformer-based models, our results on tasks such as CoLA, QQP, QNLI, and RTE exceeds DistilBERT, being close to BERT (within average 1.0 point). Overall, our model compares well to ALBERT and BERT, retaining 99\% and 98\% of the performance, respectively.

For Chinese tasks, our model gives comparable results with BERT (within 0.05 points of accuracy) and slightly better than ALBERT (78.78 vs. 77.92), which uses the same amount of training corpus. Compared with other models which apply more datasets, our model also performs well in tasks such as TNEWS and IFLYTEK. Moreover, some (e.g. RoBERTa-wwm-ext, MacBERT) inherit the parameters of BERT for continuing pre-training while our model is trained from scratch. Some baselines (e.g. BERT-wwm) uses whole word masking strategy for span masking prediction, we leave these configurations for further study.

\begin{table*}[t]
	\centering
	\small
		\scalebox{0.88}{
	\begin{tabular}{lcccccc}
        \toprule
    	 \textbf{Model} & \textbf{Settings} & \#\textbf{Param}. & \textbf{Len=64} & \textbf{Len=256}  & \textbf{Len=384}  & \textbf{Len=512} \\
    	 \midrule
    	 ELMo & 2 Bi-LSTM layer & 93M & 0.109&0.381 &0.564 & 0.745\\
    	 DistilBERT&6 encoder layers& 66M&0.016&0.021&0.025&0.034\\
         RoBERTa-base&12 encoder layers& 125M&0.017&0.026&0.042&0.051\\
         BART-base&6 encoder \& decoder layers& 140M &0.018&0.033&0.047&0.063\\
          \midrule
         \multirow{4}{*}{Ours}&6 layers, 1280 hidden size& 107M&0.010 (1.7$\times$)&0.010 (2.6$\times$)&0.010 (4.2$\times$)&0.011 (4.6$\times$)\\
         &12 layers, 1280 hidden size& 107M&0.018 (0.9$\times$)&0.018 (1.4$\times$)&0.019 (2.2$\times$)&0.019 (2.7$\times$)\\
         &6 layers, 2048 hidden size& 238M&0.011 (1.5$\times$)&0.011 (2.4$\times$)&0.012 (3.5$\times$)&0.013 (3.9$\times$)\\
         &12 layers, 2048 hidden size& 238M&0.020 (0.9$\times$)&0.020 (1.3$\times$)&0.021 (2.0$\times$)&0.021 (2.4$\times$)\\
    	\bottomrule
	\end{tabular}
	}
	\caption{Time cost (second) during inference for different architectures. Numbers in the parentheses denote the speedup compared to RoBERT-base.}
	\label{table:speed}
\end{table*}

\section{Analysis of Computational Efficiency}
\label{sec:Speed}

\textbf{Inference Speed}. We compare the inference speed of our model with different architectures, including ELMo, an RNN-based model; RoBERTa, which is based on a full Transformer encoder; DistilBERT, a lighter version of BERT with 40\% parameter reduction; and BART, which is a sequence-to-sequence model using Transformer. We test the time cost for model inference (i.e, calculating final layer representation) for a batch of sentences with different lengths. Evaluations are conducted in the same environment.

The average results over five runs are shown in Table~\ref{table:speed}. ELMo gives the lowest results as the sequential nature of RNN structure. For Transformer-based models, DistilBERT shows the minimum time cost because of the lightweight architecture, BART is slower than RoBERTa due to the nonparallel computation in the decoder. All the models take much more time when the sequence becomes much longer. For example, sequences with a length of 512 need about 3 times computational time than sequences with a length of 64.  For our model, adding the recurrent layer and the hidden size will both leads to more inference time. However, by increasing the sequence length, the inference cost grows much slower than the baselines when the sequence length reaches 256 or more, our model can give a 2$\sim$3 times speedup than DistilBERT or RoBERTa, even for large model settings. 

\begin{figure}[!t]
	\centering
	\includegraphics[scale=0.22]{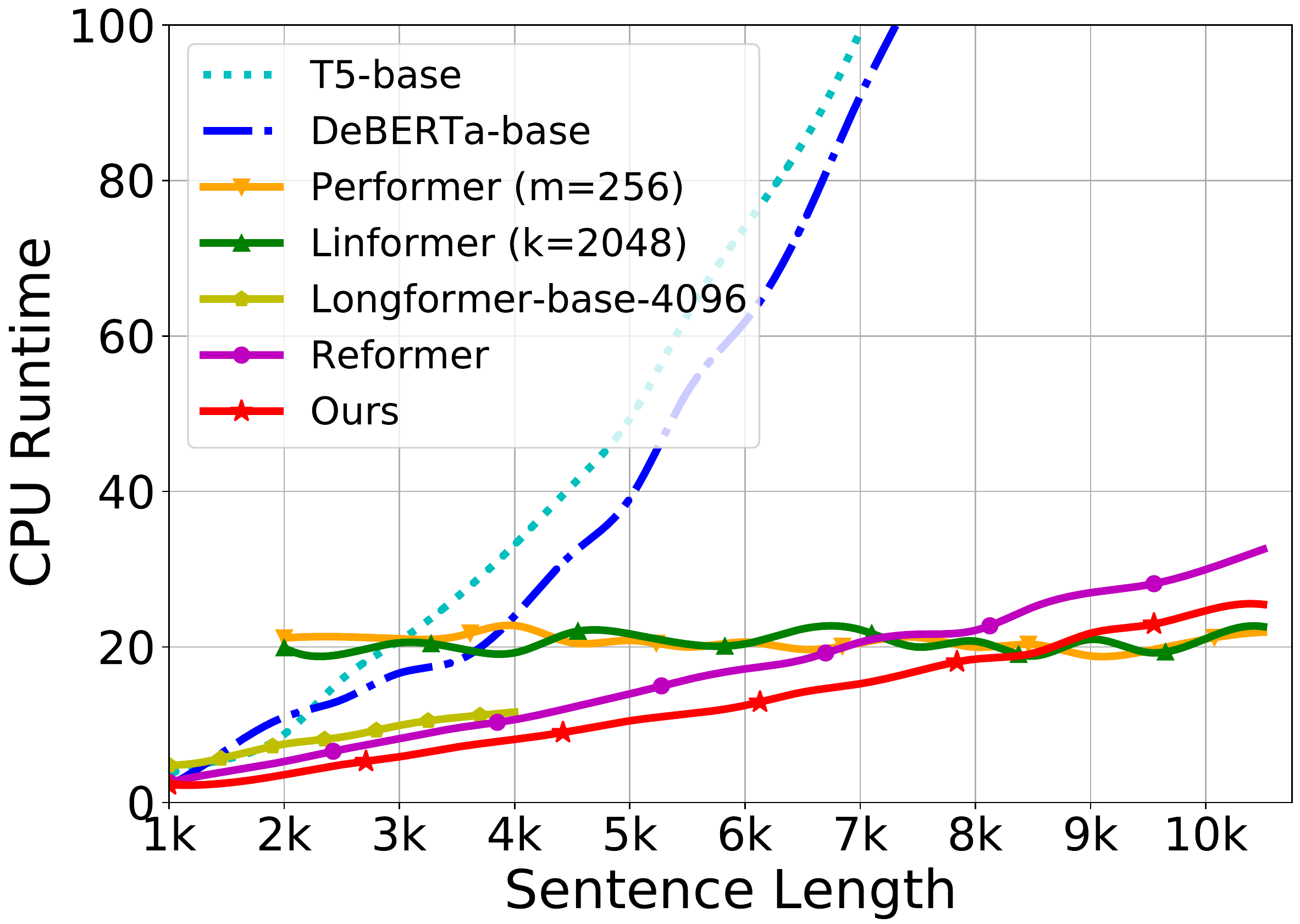}
	\caption{Comparison between models for computing long sequences.}
	\label{figure:runtime}
\end{figure}

\textbf{Extra Long Sequences}. To further study the advantages of linear complexity in our model, We compared our model with Transformer variants, including Longformer~\cite{longformer}, Reformer~\cite{reformer}, Linformer~\cite{linformer}, and Performer~\cite{performer}, which aims to optimize the self-attention mechanism for better dealing with long sequences. As standard autoencoder and seq2seq PTMs, we also include DeBERTa~\cite{deberta} and T5~\cite{t5} for reference. We use CPU runtime as the metric, avoiding the need for external resources and also enlarging the model differences, the results are shown in Figure~\ref{figure:runtime}. For DeBERTa and T5, the runtime increases quadratically as the sequence length becomes longer. The other models show the effectiveness of their improvement over the vanilla Transformer.

Longformer and Reformer give almost linear growth of runtime w.r.t sequence length. Our model is the fastest when the sequence length is below 8.5k. Linformer and Performer give slightly faster speed when the size reaches 10k. However, the models are particularly designed for long sequences. For example, Linformer project the full self-attention and find the low-rank representation, reducing the complexity from $\mathcal{O}(n^2)$ to $\mathcal{O}(nk)$, thus the projection dimension $k$ should be pre-defined and less than the sequence length $n$. Similarly, Performer pre-defined kernel feature numbers $m$ and reduce the complexity from $\mathcal{O}(n^2)$ to $\mathcal{O}(nm)$, the most computational efficiency is achieved only when $n$ is relatively large.
Overall, our model is attention-free and thus it can handle both short and long sequences friendly.

\section{Intrinsic Characteristics of Model Representation}
Since the nature of layer structure is different between S-LSTM and Transformer models, it can be an interesting research question how the characteristics of contextualized representation differ between such models. To this end, we try to investigate the intrinsic characteristics of representations. We consider doing a comparison from three aspects: 1) The space distributions of model outputs; 2) The contextualized feature redundancy of token representations in the same context; 3) The hidden states transitions in each layer.\footnote{In this section, we keep all models with the same layers and output dimensions for a fair comparison. Specifically, we re-train our model with 24 layer and 1024 hidden sizes, compared with the large version of BERT, RoBERTa, and ALBERT.}.

\subsection{Model Output Distributions}
Previous studies find that the distribution of representations from PTMs can be highly anisotropic, where the output vectors assemble in a narrow cone rather than being uniform in all directions~\cite{ethayarajh-2019-contextual,cai2020isotropy}. Such a phenomenon leads to the inferior performance of token or sentence representations~\cite{reimers-gurevych-2019-sentence,zhang-etal-2020-revisiting,DBLP:conf/aaai/ZhouL021}. To compare the output vector distribution with other models, we select 5000 random tokens from corpus, computing the cosine similarity ${\rm cos}({h_i},{h_j}) = \frac{<{h_i},{h_j}>}{\| {h_i},{h_j}\| }$ of outputs from random tokens $t_i$ and $t_j$. For sentence-level representation, we also compute the ${\rm cos}({h_i}_{[\rm{CLS}]},{h_j}_{[\rm{CLS}]})$ and ${\rm cos}({g_i},{g_j})$ for random sentences $s_i$ and $s_j$.

\begin{figure*}[!t]
	\centering
	\includegraphics[scale=0.35]{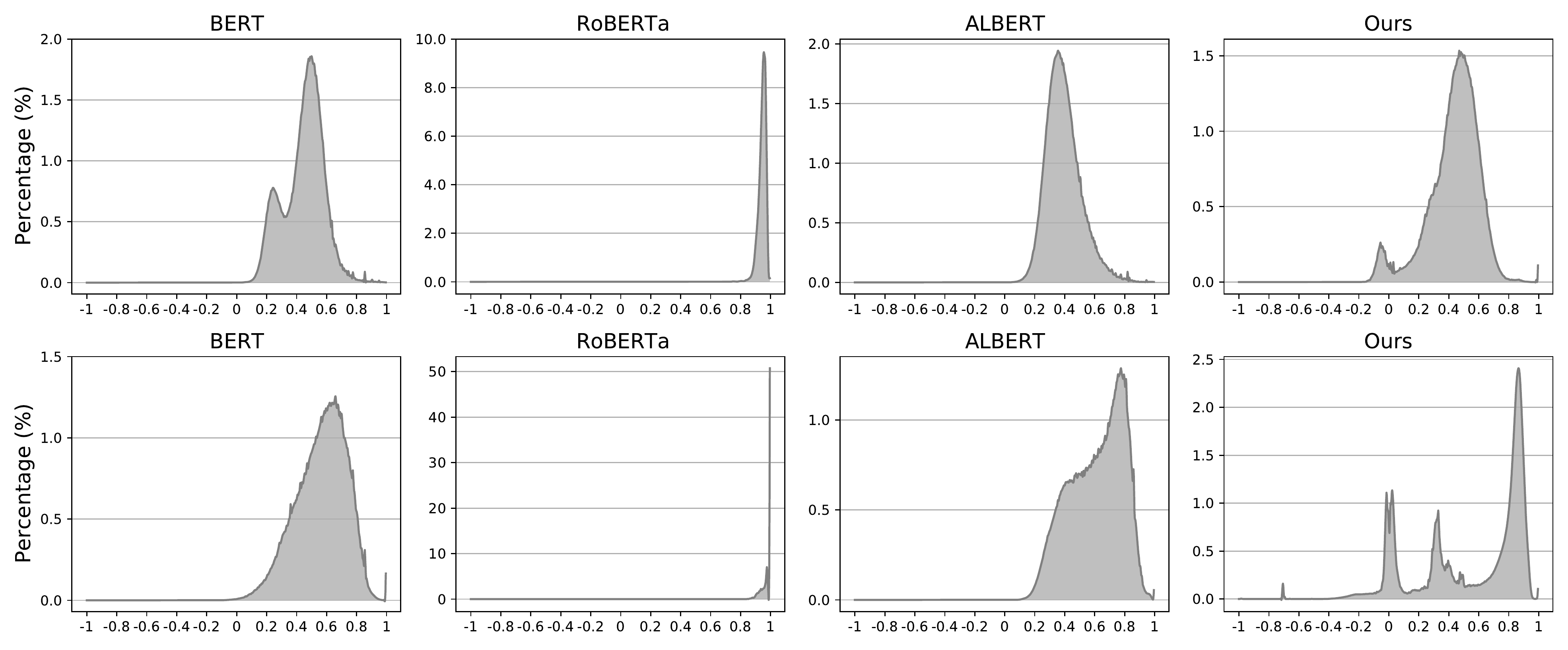}
	\caption{Cosine similarity distributions for outputs of BERT, RoBERTa, ALBERT and our model. Top: Token vectors; Bottom: Sentence vectors.}
	\label{figure:rep_1}
\end{figure*}

The results are shown in Figure~\ref{figure:rep_1}, where each figure depicts the distribution of $\binom{5000}{2}$ random similarity scores. Ideally, the similarity  distribution should be symmetric around the zero point. In the top row, the scores for BERT, RoBERTa, and ALBERT are almost all larger than zeros, where the distributions is extremely centralized around 0.8$\sim$1 for RoBERTa, this is in line with Gao et.al.~\cite{DBLP:conf/iclr/GaoHTQWL19}, where they find any two words are highly positively correlated. BERT, ALBERT and ours demonstrate similar and wider distributions, however, we find that there exist a mount of word pairs with negative relationships in our model.

As for sentence representations, BERT and ALBERT use next sentence prediction and sentence order prediction during pre-training, while RoBERTa does not introduce any sentence-level task.
However, compared to the token vectors, the distributions for sentence representation are all become more closer to 1. This may due to the [CLS] token in Transformer-based models are treated equally with other tokens, with a fixed position embeddings for zeros, thus the shape of distribution are generally similar while the correlations become more positive. The results of our model show distinct patterns, where some sentence pairs give very small similarity with -0.7, and there are also part of disjunctive areas around -0.1$\sim$0.1, 0.2$\sim$0.4 and 0.7$\sim$0.9.

Some work suggest that the high anisotropy is inherent to, or least a by-product of contextualization~\cite{ethayarajh-2019-contextual,rajaee-pilehvar-2021-cluster}, Gao et al.~\cite{DBLP:conf/iclr/GaoHTQWL19} attribute this to the nonuniform word frequencies in training data. Our results show that this could also related to the architecture designing, where the hierarchical local information and decoupled sentence representation can mitigate such phenomenon.

\subsection{Contextualized Feature Redundancy}
In the previous section, we consider the similarity of random token and sentence embedding, which reflect the general distribution in output space. In this section, we further consider the tokens in the same context, where all tokens share the same contextualized information. Existing work shows that, for Transformer-based models, each token attends to all other tokens with an attention mechanism in every layer, the overlapped context information may lead to feature redundancy. For example, Clark et al.~\cite{clark-etal-2019-bert} and Kovaleva et al.~\cite{kovaleva-etal-2019-revealing} find many attention heads generate similar attention matrices in the same context. Peng et al.~\cite{peng-etal-2020-mixture} find attention heads can be largely pruned. Dalvi et al.~\cite{dalvi-etal-2020-analyzing} suggest that 85\% of the neurons across the network can be removed without loss of accuracy.

We use principal  component analysis to investigate such contextualized feature redundancy. Formally, given sentence $s$ = $w_1,w_2,...,w_n$, we collect the feature matrix $H \in R^{n\times d}$, where each row $h_i\in R^{1\times d}$ represent the output of tokens $w_i$. By truncated singular value decomposition, the feature matrix can be written as:
\begin{equation}
H^{n\times d}=U^{n \times n}\Sigma^{n \times d}V^{d \times d} \approx  U^{n \times k}\Sigma^{k \times k}V^{k \times d}
\label{eq:svd}
\end{equation}
where $U$ and $V$ are orthogonal matrices, $\Sigma$ is the diagonal matrix with sorted singular values $\sigma_1$, ... , $\sigma_n$. The number of principal components $k$ can be selected according to the required retained information, or the Frobenius norm of the difference between the original feature matrix and the reconstructed matrix.

\begin{table}[t]
	\centering
	\small
	\begin{tabular}{ccccccc}
        \toprule
    	 \textbf{Length} &\textbf{Model} & \textbf{90\%} & \textbf{92\%} & \textbf{94\%}  & \textbf{96\%}  & \textbf{98\%} \\
    	 \midrule
    	 \multirow{4}*{$n$=100}&BERT& 48.98&53.90&59.98&67.70&79.34\\
    	 &RoBERTa& 50.50&55.24&61.06&68.62&79.54\\
    	 &ALBERT& 51.26&56.18&62.20&69.96&82.30\\
    	 &Ours& \textbf{52.56}&\textbf{58.42}&\textbf{65.58}&\textbf{74.58}&\textbf{87.18}\\
    	 \midrule
    	 \multirow{4}*{$n$=300}&BERT& 103.9 &118.6 &137.8 & 165.1& 212.6\\
    	 &RoBERTa& 109.7 & 124.3&143.1 &169.4 & 214.2\\
    	 &ALBERT&102.7  &117.5 &137.0 &165.7 &217.7 \\
    	 &Ours&\textbf{119.3}&\textbf{136.4}&\textbf{158.1}&\textbf{187.9}&\textbf{237.0}\\
    	\bottomrule
	\end{tabular}
	\caption{Principal component of features ($k$ in Eq.~\ref{eq:svd}) according to different amount of information retained. Averaged results over different sentences are reported.}
	\label{table:rep_2}
\end{table}

\begin{figure}[t]
	\centering
	\includegraphics[scale=0.42]{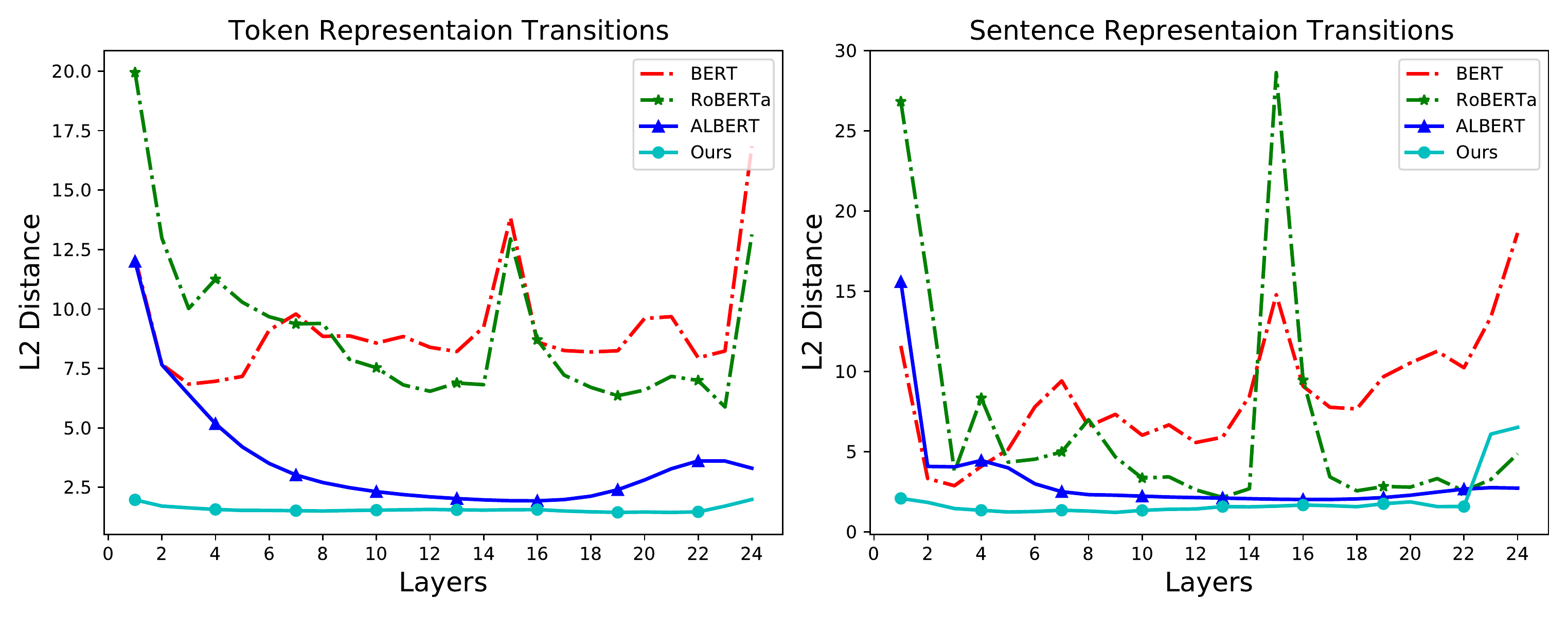}
	\caption{Hidden state transitions from layer to layer. Left: Token representations. Right: Sentence representations.}
	\label{figure:rep_3}
\end{figure}

Table~\ref{table:rep_2} shows the results. For BERT, RoBERTa and ALBERT, about 50/80 principal  components keep the 90\%/98\% information when the sequence length is 100. The numbers are around 110/210 when the length increase to 300. Our model does not use the full multi-head attention mechanism to build the contextualized information once. Instead, by combining explicit hierarchical local information and gated global information, each token receives different levels of information in each layer. In particular, our model gives the largest values accordingly (52/87 and 119/237), showing that the token representations share less common features than Transformer-based models.

\subsection{Hidden State Variance Across Layers}

The recurrent state transitions in our model improve the parameter efficiency. By sharing the parameters, the interactions between layers can be different from models without such limitations. We are interested in the layer transitions in the intermediate layers for recurrent (ALBERT, ours) and non-recurrent models (BERT, RoBERTa).

Figure~\ref{figure:rep_3} shows the L2 distances of the input and output embeddings in each layer. For both token and sentence embeddings, the transitions in BERT and RoBERTa are oscillating, where the parameters in each layer is more flexible. Results of ALBERT and our model are smoother, except for the increase of distance changes in the two-end layers, showing that
weight-sharing has an effect on stabilizing network parameters~\cite{albert}. Our model gets the smallest distance changes, however, they also do not converge to zero at all times. These findings show that the solutions for recurrent-style models are quite different of that from parameter flexible ones.

\section{Conclusion and Future Work}
We investigated a graph recurrent network for large-scale language model pre-training. Our model does not rely on the self-attention mechanism in Transformer, retaining linear computational complexity with respect to the sequence length. Results on language modeling and downstream tasks in both English and Chinese languages show that the inference time can be largely reduced while without much accuracy loss. Also, our model outputs show more diversity and less feature redundancy than Transformer-based ones. For future work, we will study our model for seq2seq-style pre-training as in BART or T5, exploring the applications to generation tasks such as machine translation.

\bibliographystyle{unsrt}
\bibliography{slstmbib}

\end{document}